\documentclass{article}
\usepackage{spconf,amsmath,graphicx}
\usepackage{mathrsfs}
\usepackage{amsfonts}
\usepackage{amssymb}

\usepackage{booktabs}
\usepackage{threeparttable}
\usepackage{multirow}
\usepackage{array}
\usepackage{booktabs}
\usepackage{subfigure}
\usepackage{color,xcolor}
\usepackage[backref]{hyperref} 
\hypersetup{hypertex=true,
            colorlinks=true,
            citecolor=blue
            }

\title{G\lowercase{reedy} O\lowercase{ffset}-G\lowercase{uided} K\lowercase{eypoint} G\lowercase{rouping} \lowercase{for} H\lowercase{uman} P\lowercase{ose} E\lowercase{stimation}}
\name{Jia Li$^{\dagger}$   \quad Linhua Xiang$^{\dagger}$ 
 \quad Jiwei Chen$^{\dagger}$$^{\ddagger}$  
    \quad Zengfu Wang$^{\dagger}$$^{\ddagger}$
}
\address{
    $^{\dagger}$ Department of Automation, 
    University of Science and Technology of China, 
    Hefei, China \\
    $^{\ddagger}$ Institute of Intelligent Machines, Chinese Academy of Sciences, Hefei, China \\
\small \texttt{\{jialee, xlh1995,  cjwbdw6\}@mail.ustc.edu.cn, 
    zfwang@ustc.edu.cn}
}

\begin{document}
%
\maketitle
\begin{abstract}
We propose a simple yet reliable bottom-up approach with a good trade-off between accuracy and efficiency for the problem of multi-person pose estimation.  Given an image, we employ an Hourglass Network to infer all the keypoints from different persons indiscriminately as well as the guiding offsets connecting the adjacent keypoints belonging to the same persons. Then, we greedily group the candidate keypoints into multiple human poses (if any), utilizing the predicted guiding offsets. And we refer to this process as \textit{greedy offset-guided keypoint grouping} (GOG). Moreover, we revisit the encoding-decoding method for the multi-person keypoint coordinates and reveal some important facts affecting accuracy. 
Experiments have demonstrated the obvious performance improvements brought by the introduced components. Our approach is comparable to the state of the art on the challenging COCO dataset under fair conditions.  The source code and our pre-trained model are publicly available online\footnote{\textcolor{magenta}{\url{https://github.com/hellojialee/OffsetGuided}}}.

\end{abstract}
\section{Introduction}
\label{sec:intro}

The problem of multi-person pose estimation is to localize the 2D skeleton keypoints (body joints) for all persons given a single image  \cite{Ronchi2017Benchmarking, Papandreou2018PersonLab, li2020simple}. It is a fundamental task with many applications.
Some recent work can address the problem of single-person pose estimation extremely well thanks to the development of  convolutional neural networks (CNNs) specially designed for human pose estimation such as Hourglass Network \cite{Newell2016Stacked, law2018cornernet, li2020simple} and HRNet \cite{sun2019deep}. But the challenge becomes much more difficult when multiple persons appear in the scene at the same time. And it has not been solved well so far considering accuracy, speed and simplicity.

\textbf{Approach taxonomy.} Existing approaches for this problem can be roughly divided into two categories: \textbf{\textit{top-down}} and \textbf{\textit{bottom-up}}. Some recent top-down approaches \cite{zhang2020distribution, Chen2017Cascaded, Papandreou2017Towards, sun2019deep, He2017Mask} have achieved high accuracy. However, most of them have low prediction efficiency especially when many people exist in the scene, for they must rely on an advanced person detector to detect all the persons and estimate all the single-person poses one by one. The bottom-up approaches  \cite{Cao2017Realtime, Newell2017Associative, kreiss2019pifpaf, li2020simple, cheng2019higherhrnet} instead infer the keypoint positions and corresponding grouping cues of all persons indiscriminately using a feed-forward network (here, we mainly discuss CNN-based approaches) and then group the detected keypoints into individual human poses. The run time of the grouping process may be very fast and nearly constant regardless of the person number in the image.
Therefore, developing bottom-up approaches is challenging but attractive for a better trade-off between accuracy and efficiency. In this paper, we mainly focus on the bottom-up approaches.

\textbf{Keypoint coordinate encoding.} Here, we only review the most important work. With the arrival of the deep learning era, DeepPose \cite{Toshev2013DeepPose} for the first time uses  CNNs to regress the Cartesian coordinates of a fixed number of person keypoints directly. As a result, this approach cannot handle the situation of multiple persons.  PersonLab  \cite{Papandreou2018PersonLab}, CenterNet \cite{Zhou:2019ta} and PifPaf \cite{kreiss2019pifpaf} decompose the task of keypoint localization into two subproblems at each position: binary classification to discriminate whether this current pixel is a keypoint, and offset regression to the ground-truth keypoint position. However, special techniques are essential to make their approaches work well, e.g., leveraging focal loss \cite{Lin2017Focal} for the classification task and Laplace-based L1 loss \cite{kendall2017uncertainties} for the offset regression task \cite{kreiss2019pifpaf}. By contrast, Tompson  et al. \cite{Tompson2014Joint} and many later researchers use CNNs to predict the Gaussian response heatmaps of person keypoints and then obtain the keypoint coordinates by finding the local maximum responses in the heatmaps. But high-res feature maps are required to alleviate the precision decline of keypoint localization.

\textbf{Keypoint grouping encoding.}
The encoding of keypoint grouping (or association) information is critical for the post-processing in bottom-up approaches. 
Here, we conclude the existing approaches for keypoint grouping into two categories: \textbf{\textit{global grouping}} and \textbf{\textit{greedy grouping}}. Prior work, such as DeeperCut \cite{Insafutdinov2016DeeperCut}, Associative Embedding \cite{Newell2017Associative} and CenterNet \cite{Zhou:2019ta}, presents different approaches to encode the global keypoint grouping information. For instance, Associative Embedding employs Hourglass Networks \cite{Newell2016Stacked} to infer the identity tags for all candidate keypoints. These tags are grouping cues to cluster keypoints into individual human poses. And CenterNet \cite{Zhou:2019ta} proposes center offsets away from person centers as keypoint grouping cues. By contrast, Part Affinity Fields \cite{Cao2017Realtime}, PersonLab \cite{Papandreou2018PersonLab} and PIFPAF \cite{kreiss2019pifpaf} are greedy grouping approaches, associating adjacent nodes in a human skeleton tree independently rather than finding the global solution of a global graph matching problem.

Some state-of-the-art (SOTA) bottom-up approaches are remarkable in estimation accuracy but have so complicated structure and many hyper-parameters affecting results that we cannot clearly figure out the contributions of introduced components or compare these approaches equally. 
In this work, we propose a bottom-up approach, which is easy to follow and works well in terms of accuracy, speed and clarity. We use Hourglass-104 \cite{law2018cornernet} as the inference model and select CenterNet \cite{Zhou:2019ta} as the baseline approach.  The main contributions of this paper are as follows:

(1) We present a greedy keypoint grouping method, which we refer to as \textit{greedy offset-guided keypoint grouping} (GOG). The adjacent keypoints of each person are connected by guiding offsets. The proposed GOG algorithm is robust and fast.

(2) We follow and improve the existing Gaussian heatmap encoding-decoding method for keypoint coordinates of the multi-person poses, ensuring precise keypoint localization.

(3) Our preliminary system has obtained obvious performance increases compared with the baseline.

\section{METHOD}
\label{sec:Section2}

\subsection{Preliminary}

\begin{figure}[!tb]
	\centering
	\includegraphics[width=8.2cm]{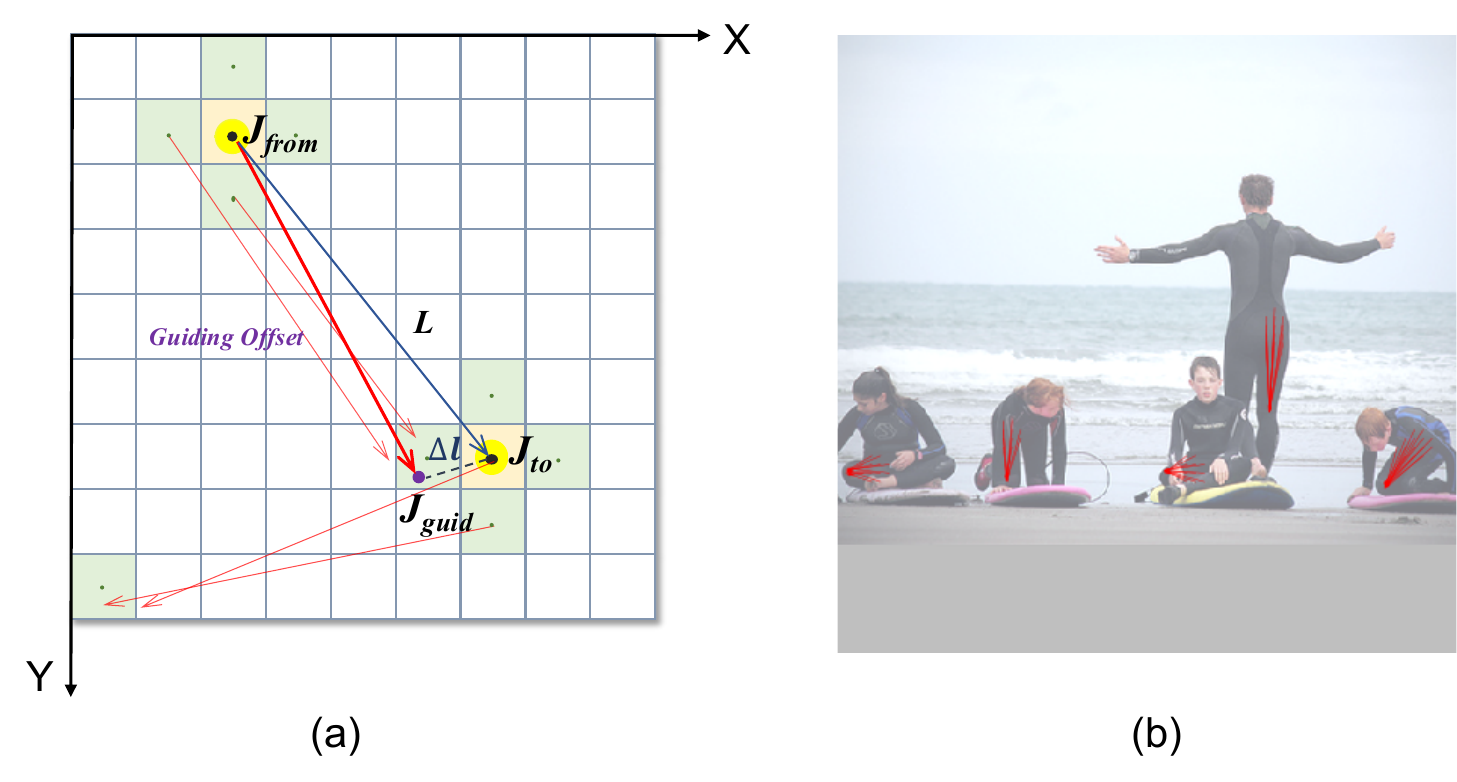}
	\vspace{-4mm}
	\caption{Definition of guiding offsets. (a) The \textcolor{blue}{ground-truth} guiding offset placed at the position of keypoint $J_{from}$ points to its adjacent keypoint $J_{to}$. Actually, $J_{from}$ and $J_{to}$ are in different heatmap channels, the figure is just for description. In practice, the \textcolor{red}{inferred} guiding offset points to the floating-point position $J_{guid}$. (b) The inferred guiding offsets at the area around the ``right hip" keypoints. They guide to the corresponding ``right ankle" keypoints of the same individuals.}
	\label{fig:guidingoff}
	\vspace{-4mm}
\end{figure}

\begin{figure*}[!htb]
	\centering
	\includegraphics[width=13.5cm]{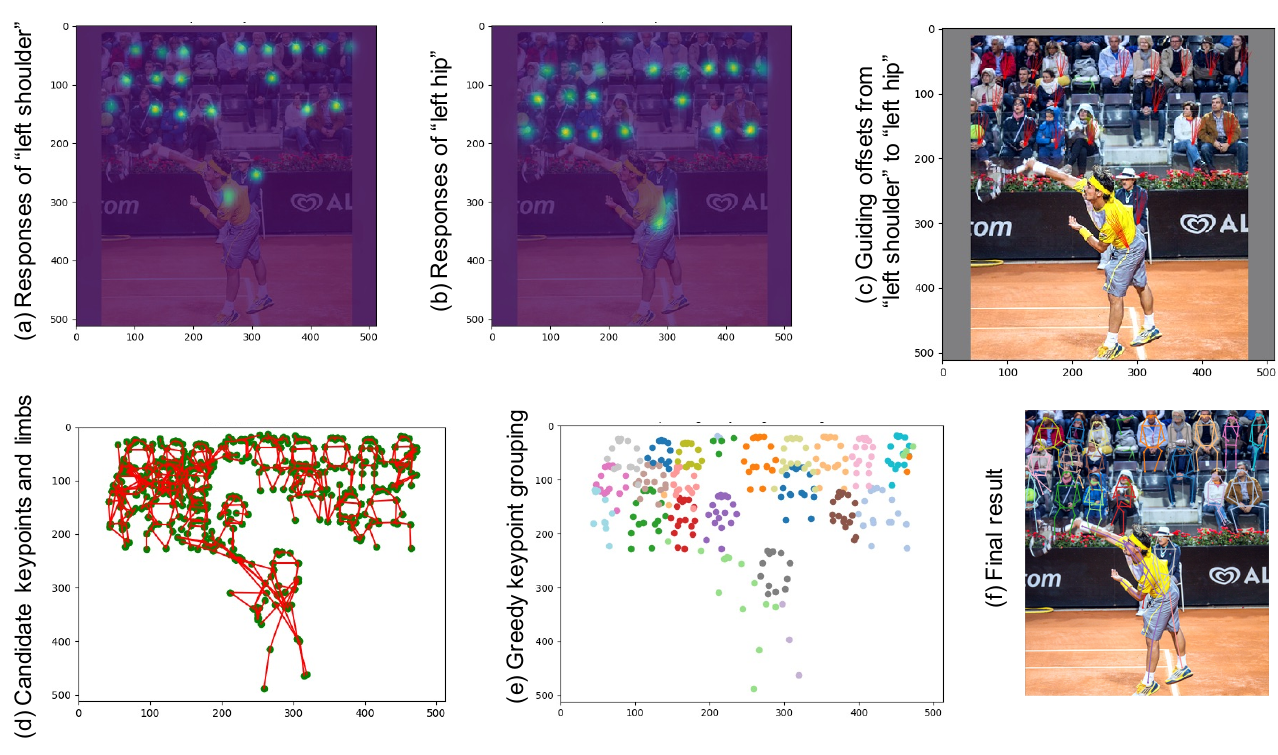}
	\vspace{-4mm}
	\caption{Overview of the proposed approach. \textbf{Top-left}, \textbf{top-middle} and \textbf{top-right}: we infer the keypoints and the guiding offsets simultaneously using Hourglass-104 \cite{Newell2016Stacked, law2018cornernet, Zhou:2019ta}. There are 17 types of keypoint heatmaps and 19 types of guiding-offset feature maps. \textbf{Bottom-left}: next, the top $k$ ($k=32$ in all our experiments) scoring candidate keypoints and limbs (connections of paired keypoints) of each type and with high confidence are collected. \textbf{Bottom-middle}: subsequently, they are grouped into individual human poses greedily using our GOG algorithm. \textbf{Bottom-right}: finally we measure the confidence of each human pose by averaging the response values of its keypoints and filter out the human poses with low confidence.}
	\label{fig:pipeline}
	\vspace{-4mm}
\end{figure*}

Keypoint grouping information is essential for multi-person pose estimation. Here, we propose a novel grouping encoding for keypoints as the form of \textit{guiding offsets} to greedily ``connect'' the adjacent keypoints belonging to the same persons. A guiding offset illustrated in Fig.\ref{fig:guidingoff} is the displacement vector starting at a keypoint $J_{from}$ of a person and pointing to his/her next keypoint $J_{to}$. 
An overview of our approach is shown in Fig.\ref{fig:pipeline}, which is mainly inspired by Simple Pose \cite{li2020simple} and CenterNet \cite{Zhou:2019ta} but different from them or other bottom-up approaches \cite{Papandreou2018PersonLab, kreiss2019pifpaf, cheng2019higherhrnet, Newell2017Associative} based on offsets.

PersonLab \cite{Papandreou2018PersonLab} and PIFPAF \cite{kreiss2019pifpaf} detect an arbitrary root keypoint of a person and use bidirectional offsets to generate a tree-structured kinematic graph of that person. Merely the root keypoint coordinate is obtained in the confidence maps of classification while the subsequent keypoint coordinate is regressed directly on the basis of the offset fields and the former keypoint location. Abundant root keypoints are detected as the seeds of the kinematic graphs to ensure they can detect as many human poses as possible. Hence, they have to apply non-maximum suppression (NMS) to remove redundant person poses. 
CenterNet \cite{Zhou:2019ta} detects the center point of each person as the root ``node" and localizes keypoints using classification heatmaps and refinement offsets. CenterNet only associates the center point with the keypoints which are within the same person bounding box using center offsets. 

However, both precise offset regression to keypiont locations and localization of person center points are vague visual tasks. Previous CNN-based work such as DeepPose \cite{Toshev2013DeepPose} and Tompson et al.  \cite{Tompson2014Joint} has already suggested that inferring keypiont heatmaps is easier than inferring keypoint Cartesian coordinates. The vague ground-truth keypoint of a person is different from his/her bounding box or segmentation boundary with clear visual concepts. As a result, we localize all keypoints using Gaussian response heatmaps. As for keypoint grouping, we infer a set of guiding offsets at the positions close to every keypoint pointing to an adjacent keypoint.

\subsection{Definition of Gaussian Responses}
\label{section: gaussin heatmap}
We use the most popular coordinate encoding-decoding method for keypoint localization, i.e., Gaussian response heatmaps. Assuming $I \in {\mathbb{R}}^{W \times H \times 3}$ is the input image of width $W$ and height $H$, the ground-truth keypoint heatmap for the network output is $G^{*} \in [0,1]^{\frac{W}{R} \times \frac{H}{R} \times C}$, in which $R$ is the output stride and C is the number of keypoint types. In our approach, $R=4$ for Hourglass Networks \cite{Newell2016Stacked, li2020simple, law2018cornernet, Zhou:2019ta} and $C=17$ for COCO dataset. Considering a pixel value  at $g(u, v)$ in the $c$-th ground-truth keypoint heatmap ${\mathbb{R}}^{\frac{W}{R} \times \frac{H}{R} }$ , before generating the ground truth at $g$, we map $g$ to the position $\tilde {g} (x, y)$ in the input image space ${\mathbb{R}}^{W \times H \times C}$ using this transformation \cite{li2020simple,Cao2017Realtime}: ${\tilde {g}}(x, y) = g(u, v)\cdot R + R/2 -0.5$. If the ground-truth keypoint of type $c$ at the position ${p} \in {\mathbb{R}}^{W \times H \times C}$ is the nearest keypoint away from ${\tilde {g}}$, we generate the ground-truth Gaussian response value at $g$ as:
$
	G^{*}(g(u, v), \sigma_{k}) = 
	\exp( -\frac{({\tilde {g}}_{x}-p_{x})^{2} + ({\tilde {g}}_{y}-p_{y})^{2}}{2\sigma_{k}^2}  )
	\label{eq:guassian}
$
, where $\sigma_{k}$ is fixed for different keypoint types for simplicity. 
The coordinate transformation from $g$ to $\tilde {g}$ is critical for keypoint localization precision because we hold the perspective everywhere in this work that each pixel in the image occupies a $1 \times 1$ ``area" and that pixel value lies exactly at the center of the pixel cell.  

During decoding (testing), we correspondingly upsample the output heatmaps of the network $R$ times using bicubic interpolation and then find the local maximums in heatmaps as detected keypoints.  It should be notated that the ground-truth keypoint coordinates are integers in the COCO dataset. And in this case, original keypoint locations can be restored in this way. On the contrary, the theoretical error keypoint localization is inevitable when we use bilinear interpolation.

\subsection{Definition of Guiding Offsets}
\label{subsec:: definition of guiding offset}

As illustrated in Fig.\ref{fig:guidingoff} (a), the ground-truth guiding offset originated from keypoint $J_{from}$ to kyepoint $J_{to}$ is a displacement vector calculated as $\{ J_{to}(x) - J_{from}(x),  J_{to}(y) - J_{from}(y)\}$. We place a set of ground-truth guiding offsets around the ``start" keypoint $J_{from}$. A human pose skeleton with 19 ``limbs'' is defined same as in PIFPAF \cite{kreiss2019pifpaf}. A limb in the human skeleton is a ``connection'' between two adjacent keypoints. Consequently, we use 19 types of guiding offsets to associate keypoints unless mentioned otherwise. 

During decoding, we only use the guiding offsets predicted at the local maximum  positions in the keypoint heatmaps as the grouping cues.  Supposing we have two candidate keypoints in Fig.\ref{fig:guidingoff} (a) to be paired and they have Gaussian response values $J_{from}(s)$ and $J_{to}(s)$ respectively. We measure the connection score between them as: $S(J_{from}, J_{to}) = J_{from}(s) \cdot J_{to}(s) \cdot \exp(- \frac{\Delta l}{L})$, where $L$ is the length between $J_{from}$ and $J_{to}$, $\Delta l$ is the ``error" distance between $J_{guid}$ and $J_{to}$. Here, we refer to the connection of the paired keypoints with high connection confidence as a candidate ``limb".

\subsection{Greedy Keypoint Grouping}
We have already described the guiding offsets and the human pose skeleton. Assuming we have detected a keypoint $J_{from}$ and a keypoint $J_{to}$ of the adjacent types in Fig.\ref{fig:guidingoff} (a), they are paired and connected as a limb if the connection score $S(J_{from}, J_{to})$ is above a threshold. The adjacent keypoints in the human skeleton are paired and measured independently.
Greedy offset-guided keypoint grouping (GOG) is the process to assemble the collected limbs, i.e., the adjacent keypoint connections, into individual human skeletons as shown at the \textbf{Bottom} of Fig.\ref{fig:pipeline}.
Our GOG algorithm is inspired by Simple Pose \cite{li2020simple} and CenterNet \cite{Zhou:2019ta}, which is greedy and pretty fast.
Note that one keypoint can not be connected to two or more keypoints of the same type.  The details can be found in our source code.

\subsection{Network Structure and Losses}
Following CenterNet \cite{Zhou:2019ta}, we employ Hourglass-104 as the backbone network that consists of two stacked hourglass modules \cite{Newell2016Stacked} and has the depth of 104. We only add $1 \times 1$ convolutional layers into the backbone to regress the desired outputs. The focal $L_{2}$ loss \cite{li2020simple} is applied to supervise the regression of keypoint heatmaps while only the guiding offsets close to ``start" keypoints are supervised by the $L_{1}$ loss and normalized by individual scales.

\section{Experiments}

\begin{table}[tp]
	\centering
	\setlength{\belowcaptionskip}{-1pt} 
	\caption{
		Results of accuracy and speed on the COCO validation set. We use single-scale inference and the same computer with a 2080 Ti GPU to test speed equally. The faster speed in parenthesis is obtained using AMP FP16 inference.
	}
	\label{tab:ablation results}
	\small
	\centering
	\begin{threeparttable}
		\scalebox{0.8}{
			\begin{tabular}{p{4.2cm} | c | c | c }
				\toprule
				\bf Method & AP & FPS (FP16) & Input \\
				\midrule  
				CenterNet (\textbf{baseline}) \cite{Zhou:2019ta}, w/ flip & 64.0 &8.3  &$\sim$640 \\
				\midrule  
				Simple Pose \cite{li2020simple}, w/ IMHourglass & 66.3 &0.7  &768   \\
				\midrule  
				PersonLab \cite{Papandreou2018PersonLab}, w/ RestNet-101 &61.2  &-   & 801  \\ 
				\midrule 
				PIFPAF \cite{kreiss2019pifpaf}, w/ RestNet-101 & 65.7 &4.0  &641   \\
				\midrule 
				Ours 1, w/o flip & 64.4   & 27.8 (36.6) & 640  \\
				\midrule  
				Ours 2 (\textbf{final}), w/ flip & 66.1   & 14.7 (21.4) & 640  \\  
				\midrule  
				Ours3 (plus), w/ flip, w/ 39 limbs   & 67.5 & 7.9 (11.9)  & 768 \\
				\bottomrule
		\end{tabular}}
	\end{threeparttable}
	\vspace{-3mm}
\end{table}

\textbf{Implementation details.} Our system is implemented using Python and Pytorch.  We present experiments on the COCO  dataset.
It consists of  the training set, the test-dev set, and the validation set.
 Our models are trained on the training set and evaluated on the validation set (excluding images without people) and test-dev set. We set ${{\sigma }_{k}}$ of the Gaussian response kernel to $7$. The supervision area of the guiding offsets around the ``start" keypoint is set to $7 \times 7$.
The training images are cropped to $512 \times 512$ patches with random scaling, rotation, flipping and translation. We fine-tune the pre-trained Hourglass-104 in \cite{Zhou:2019ta} using the Adam optimizer with the batch size of 32, the initial learning rate of 1e-3 for less than 150 epochs. For evaluation we use single-scale inference, keeping the same as CenterNet \cite{Zhou:2019ta}, PersonLab \cite{Papandreou2018PersonLab}, and PIFPAF \cite{kreiss2019pifpaf} for fair comparisons.

\textbf{Ablation studies.} It is essential to learn the upper limit of our system.  We replace the network outputs with the ground-truth values and obtain the theoretical upper bound: 86.6$\%$ AP on the COCO validation set.  If we define more limbs in the human skeleton, let us say 44 limbs, to connect adjacent keypoints, the upper bound reaches 91.0$\%$ AP. It further increases to 95.5$\%$ AP when we set the output stride $R$ to 1. 

Please refer to Table \ref{tab:ablation results}, the baselin  CenterNet and all our results are based on Hourglass-104 with single-scale inference. The long side of the input image is resized to a fixed number. Our approach (Ours 2) significantly exceeds the baseline (\textbf{$+$2.1$\%$} AP) on the COCO validation set under equal conditions. And the enhanced result, Ours 3 with more limbs and a larger input, achieves \textbf{67.5$\%$} AP as well as good speed. The results in Table \ref{tab:ablation results} suggest that our approach has obtained a satisfying trade-off between accuracy and speed.

\textbf{Keypoint coordinate encoding-decoding.} We validate different encoding-decoding methods for multi-person keypoint coordinates in Table \ref{tab:encoding-decoding}, in which, ``qnt'' represents quantifying the numerical keypoint to an integer coordinate before generating the Gaussian distribution, ``ro'' is short for refinement offsets improving localization precision. Here, flip augmentation is not used.
 DARK \cite{zhang2020distribution} works pretty well in some top-down approaches, and we have modified and applied it to our system.  
 Experiments indicate that our encoding-decoding method is best in average precision. 
 DARK assumes that the predicted keypoint responses obey a Gaussian distribution, but the network outputs do not meet this assumption in bottom-up approaches, in which inferred Gaussian responses are often far from ground-truth values. Refinement offsets failed in improving our result (\textbf{$-$0.3$\%$} AP), suggesting that naive offset regression to precise keypoint locations is not accurate enough. As expected in Section \ref{section: gaussin heatmap}, using bilinear interpolation in heatmap upsampling leads to a large accuracy drop (\textbf{$-$2.1$\%$} AP).

\begin{table}[!tp]
	\centering
	\setlength{\belowcaptionskip}{-1pt} 
	\caption{
		Ablation studies of the encoding-decoding methods for keypoint coordinates on the COCO validation set.
	}
	\label{tab:encoding-decoding}
	\small
	\centering
	\begin{threeparttable}
		\scalebox{0.8}{
			\begin{tabular}{c| c | c | c | c | c | c }
				\toprule
				 &DARK \cite{zhang2020distribution} & qnt &qnt + ro &Ours &Ours + ro &Ours w/ bilinear  \\
				\midrule  
				AP &54.5 & 56.2 &62.8  &64.4 &64.1 &62.3 \\
				
				\bottomrule
		\end{tabular}}
	\end{threeparttable}
	\vspace{-3mm}
\end{table}

\textbf{Results.} We compare our approach with the reproducible SOTA bottom-up approaches in Table \ref{tab:test-dev results}. Though our results are not always the best, our preliminary system has advantages taking into account accuracy, speed, and simplicity.

\vspace{-3mm}
\begin{table}[!tp]
	\centering
	\setlength{\belowcaptionskip}{-1pt} 
	\caption{
		Results on the COCO test-dev set. We report single-scale inference results. Entries marked with ``*'' are produced by corresponding official source code and models.
	}
	\label{tab:test-dev results}
	\small
	\centering
	\begin{threeparttable}
		\scalebox{0.85}{
			\begin{tabular}{p{4.1cm} | c | c | c | c }
				\toprule
				\bf Method & AP &AP$^{M}$	  &AP$^{L}$ &Input\\
				\midrule  
				CenterNet$^{*}$ \cite{Zhou:2019ta}, Hourglass-104 & 63.0 &58.9  &70.4 &$\sim$640 \\
				\midrule  
				Simple Pose$^{*}$ \cite{li2020simple}, IMHourglass & 64.6 &60.3  &71.6 &768   \\
				\midrule 
				PersonLab \cite{Papandreou2018PersonLab}, RestNet-101 &65.5  &61.3   & 71.5 &1401  \\ 
				\midrule 
				PIFPAF$^{*}$ \cite{kreiss2019pifpaf}, RestNet-101 & 64.9 &60.6  &71.2 &641   \\
				\midrule 
				Bottom-up HRNet-W32 \cite{cheng2019higherhrnet} & 64.1 &57.4  &73.9 &$>$640   \\
				\midrule 
				Ours (\textbf{final}) & 64.7   & 60.7 &70.4  & 640  \\
			\midrule 
			Ours (\textbf{final}) & 65.6   & 63.3 &68.8  & 768  \\
				\bottomrule
		\end{tabular}}
	\end{threeparttable}
	\vspace{-3mm}
\end{table}

\section{Conclusion}
We have presented a bottom-up approach for the problem of multi-person pose estimation, which is simple enough yet achieves a respectable trade-off between accuracy and efficiency. Specifically, we have revisited and improved the encoding-decoding method for multi-person keypoint coordinates in heatmaps, leading to more precise localization.  And we have proposed the guiding offsets between adjacent keypoints as the keypoint grouping cues and assemble human skeletons greedily. Experiments on the challenging COCO dataset have demonstrated the advantages and much room for accuracy improvement of our preliminary system.
\bibliographystyle{IEEEbib}
\bibliography{refs}

\begin{thebibliography}{10}

\bibitem{Ronchi2017Benchmarking}
Matteo~Ruggero Ronchi and Pietro Perona,
\newblock ``Benchmarking and error diagnosis in multi-instance pose
  estimation,''
\newblock in {\em 2017 IEEE International Conference on Computer Vision}. IEEE,
  2017, pp. 369--378.

\bibitem{Papandreou2018PersonLab}
George Papandreou, Tyler Zhu, Liang-Chieh Chen, Spyros Gidaris, Jonathan
  Tompson, and Kevin Murphy,
\newblock ``Personlab: Person pose estimation and instance segmentation with a
  bottom-up, part-based, geometric embedding model,''
\newblock in {\em The European Conference on Computer Vision}, September 2018.

\bibitem{li2020simple}
Jia Li, Wen Su, and Zengfu Wang,
\newblock ``Simple pose: Rethinking and improving a bottom-up approach for
  multi-person pose estimation,''
\newblock in {\em Proceedings of the AAAI conference on artificial
  intelligence}, 2020, vol.~34, pp. 11354--11361.

\bibitem{Newell2016Stacked}
Alejandro Newell, Kaiyu Yang, and Jia Deng,
\newblock ``Stacked hourglass networks for human pose estimation,''
\newblock in {\em European Conference on Computer Vision}. Springer, 2016, pp.
  483--499.

\bibitem{law2018cornernet}
Hei Law and Jia Deng,
\newblock ``Cornernet: Detecting objects as paired keypoints,''
\newblock in {\em Proceedings of the European Conference on Computer Vision},
  2018, pp. 734--750.

\bibitem{sun2019deep}
Ke~Sun, Bin Xiao, Dong Liu, and Jingdong Wang,
\newblock ``Deep high-resolution representation learning for human pose
  estimation,''
\newblock in {\em Proceedings of the IEEE Conference on Computer Vision and
  Pattern Recognition}, 2019, pp. 5693--5703.

\bibitem{zhang2020distribution}
Feng Zhang, Xiatian Zhu, Hanbin Dai, Mao Ye, and Ce~Zhu,
\newblock ``Distribution-aware coordinate representation for human pose
  estimation,''
\newblock in {\em Proceedings of the IEEE/CVF Conference on Computer Vision and
  Pattern Recognition}, 2020, pp. 7093--7102.

\bibitem{Chen2017Cascaded}
Yilun Chen, Zhicheng Wang, Yuxiang Peng, Zhiqiang Zhang, Gang Yu, and Jian Sun,
\newblock ``Cascaded pyramid network for multi-person pose estimation,''
\newblock in {\em IEEE Conference on Computer Vision and Pattern Recognition}.
  IEEE, 2018, pp. 7103--7112.

\bibitem{Papandreou2017Towards}
George Papandreou, Tyler Zhu, Nori Kanazawa, Alexander Toshev, Jonathan
  Tompson, Chris Bregler, and Kevin Murphy,
\newblock ``Towards accurate multi-person pose estimation in the wild,''
\newblock in {\em 2017 IEEE Conference on Computer Vision and Pattern
  Recognition}. IEEE, 2017, pp. 3711--3719.

\bibitem{He2017Mask}
Kaiming He, Georgia Gkioxari, Piotr Doll{\'a}r, and Ross Girshick,
\newblock ``Mask r-cnn,''
\newblock in {\em 2017 IEEE International Conference on Computer Vision}. IEEE,
  2017, pp. 2980--2988.

\bibitem{Cao2017Realtime}
Zhe Cao, Tomas Simon, Shih~En Wei, and Yaser Sheikh,
\newblock ``Realtime multi-person 2d pose estimation using part affinity
  fields,''
\newblock in {\em IEEE Conference on Computer Vision and Pattern Recognition},
  2017, pp. 1302--1310.

\bibitem{Newell2017Associative}
Alejandro Newell, Zhiao Huang, and Jia Deng,
\newblock ``Associative embedding: End-to-end learning for joint detection and
  grouping,''
\newblock in {\em Advances in Neural Information Processing Systems}, 2017, pp.
  2277--2287.

\bibitem{kreiss2019pifpaf}
Sven Kreiss, Lorenzo Bertoni, and Alexandre Alahi,
\newblock ``Pifpaf: Composite fields for human pose estimation,''
\newblock in {\em Proceedings of the IEEE Conference on Computer Vision and
  Pattern Recognition}, 2019, pp. 11977--11986.

\bibitem{cheng2019higherhrnet}
Bowen Cheng, Bin Xiao, Jingdong Wang, Honghui Shi, Thomas~S Huang, and Lei
  Zhang,
\newblock ``Higherhrnet: Scale-aware representation learning for bottom-up
  human pose estimation,''
\newblock in {\em Proceedings of the IEEE Conference on Computer Vision and
  Pattern Recognition}, 2020, pp. 5386--5395.

\bibitem{Toshev2013DeepPose}
Alexander Toshev and Christian Szegedy,
\newblock ``Deeppose: Human pose estimation via deep neural networks,''
\newblock in {\em Proceedings of the IEEE Conference on Computer Vision and
  Pattern Recognition}, 2014, pp. 1653--1660.

\bibitem{Zhou:2019ta}
Xingyi Zhou, Dequan Wang, and Philipp Kr{\"a}henb{\"u}hl,
\newblock ``Objects as points,''
\newblock in {\em arXiv preprint arXiv: \\ 1904.07850}, 2019.

\bibitem{Lin2017Focal}
Tsung-Yi Lin, Priya Goyal, Ross Girshick, Kaiming He, and Piotr Dollar,
\newblock ``Focal loss for dense object detection,''
\newblock in {\em 2017 IEEE International Conference on Computer Vision}. IEEE,
  2017, pp. 2999--3007.

\bibitem{kendall2017uncertainties}
Alex Kendall and Yarin Gal,
\newblock ``What uncertainties do we need in bayesian deep learning for
  computer vision?,''
\newblock in {\em Advances in neural information processing systems}, 2017, pp.
  5574--5584.

\bibitem{Tompson2014Joint}
Jonathan Tompson, Arjun Jain, Yann LeCun, and Christoph Bregler,
\newblock ``Joint training of a convolutional network and a graphical model for
  human pose estimation,''
\newblock in {\em Proceedings of the 27th International Conference on Neural
  Information Processing Systems-Volume 1}. MIT Press, 2014, pp. 1799--1807.

\bibitem{Insafutdinov2016DeeperCut}
Eldar Insafutdinov, Leonid Pishchulin, Bjoern Andres, Mykhaylo Andriluka, and
  Bernt Schiele,
\newblock ``Deepercut: A deeper, stronger, and faster multi-person pose
  estimation model,''
\newblock in {\em European Conference on Computer Vision}. Springer, 2016, pp.
  34--50.

\end{thebibliography}

\end{document}